# A modelOps-based framework for intelligent medical knowledge extraction

Hongxin Ding
*School of Computer Science*
Peking University
Peking, China
dinghx@pku.edu.cn

Peinie Zou
*School of Software & Microelectronics*
Peking University
Peking, China
zplkq@stu.pku.edu.cn

Zhiyuan Wang
*School of Software & Microelectronics*
Peking University
Peking, China
2101210407@stu.pku.edu.cn

Junfeng Zhao
*School of Computer Science*
Peking University
Peking, China
zhaojf@pku.edu.cn

Yasha Wang
*School of Computer Science*
Peking University
Peking, China
Wangyasha@pku.edu.cn

Qiang Zhou
*Shenzhen Emergency Medical Center*
Shenzhen, China
zq_shenzhen2023@163.com

*Abstract*—Extracting medical knowledge from healthcare texts enhances downstream tasks like medical knowledge graph construction and clinical decision-making. However, the construction and application of knowledge extraction models lack automation, reusability and unified management, leading to inefficiencies for researchers and high barriers for non-AI experts such as doctors, to utilize knowledge extraction. To address these issues, we propose a ModelOps-based intelligent medical knowledge extraction framework that offers a low-code system for model selection, training, evaluation and optimization. Specifically, the framework includes a dataset abstraction mechanism based on multi-layer callback functions, a reusable model training, monitoring and management mechanism. We also propose a model recommendation method based on dataset similarity, which helps users quickly find potentially suitable models for a given dataset. Our framework provides convenience for researchers to develop models and simplifies model access for non-AI experts such as doctors.

## I. Introduction

With the development of information technology, a large amount of unstructured data has been accumulated in the medical field, including electronic medical records, medical encyclopedias, textbooks, and online consultations. To fully utilize the medical knowledge contained in these texts to assist healthcare applications, efficient and automated knowledge extraction becomes imperative. Knowledge extraction aims to extract structured factual knowledge from unstructured text, including tasks like named entity recognition and relation extraction.

In recent years, knowledge extraction models based on deep learning have been developed, exhibiting remarkable performance, such as NESSMa in the medical field [1] and BioFLAIR in the biological field [2]. However, there are still challenges in the construction and application of knowledge extraction models. During data preprocessing, excessive time is often invested in coding to adapt datasets to models. During model training and evaluation, manual intervention is needed throughout the process, from resource allocation to results tracking, making the process cumbersome and impracticable for non-AI experts. During model application, users are often faced with numerous models and can't quickly and accurately find a suitable model that meets specific requirements. These challenges result in low research efficiency and high application barrier for Non-AI experts like doctors.

Model operations (ModelOps) has been proposed to tackle the lack of scientific management of AI models and to enhance their development and application efficiency. ModelOps aims to govern the lifecycle of AI models, and enable high-quality, efficient, and sustainable AI processes. However, existing ModelOps systems mainly support classification tasks on tabular data, with limited capability for knowledge extraction task on unstructured textual data. Aforementioned issues concerning knowledge extraction models are marginally studied. Therefore, we propose a ModelOps-based Intelligent knowledge Extraction Framework, to simplify the construction and application of knowledge extraction models, offer low-code and convenient model services for users, and help expand the real-world application of medical knowledge extraction. The main contributions of this paper include:

- Design and implement a dataset abstraction mechanism based on multi-layer callback functions, enabling automated dataset adaptation. Our system integrates 12 knowledge extraction datasets and 8 models.

- Design and implement a reusable model training, monitoring and management mechanism, which automates the process of model experimentation.

- Propose a model recommendation method based on dataset similarity to address the challenge of difficult model selection. Experimental results demonstrate the effectiveness of this method.

- Design and implement a ModelOps-based medical knowledge extraction platform based on the proposed framework, allowing users to conveniently utilize knowledge extraction models in a low-code manner.

The remaining structure of this paper is as follows: Section 2 introduces the research and technologies related to our work. Section 3 describes the details of the architecture of our framework. Section 4 introduces the proposed model recommendation method based on dataset similarity and validates its effectiveness through experiments. Section 5 presents the design and implementation of the developed platform. Section 6 concludes the work and provides prospects.

## II. RELATED WORK

### A. Knowledge Extraction

Knowledge extraction has been widely used in various fields. In the medical field, it mines valuable information including entities such as diseases, medications, treatment plans, and their relations from large-scale unstructured medical texts. It provides support for medical question answering, medical knowledge graph construction, clinical decision-making and other tasks, thus enhancing medical knowledge understanding and application.

Knowledge extraction mainly includes two tasks: named entity recognition (NER) and relation extraction (RE). NER aims to recognize entities from texts and determine their types. NER datasets may use sequence labeling methods like BIOS, or directly list entities. Their file formats include txt, CSV, JSON, etc. The annotation and storage formats are diverse, without a unified standard, complicating data processing. NER methods include traditional rule-based methods, dictionary-based methods, machine learning methods, and recent deep learning methods. For instance, the BILSTM-CRF model [3] uses a Bi-LSTM network to capture text features and outputs the label sequence through a conditional random field. Li et al [4] utilize a pointer network for entity boundary prediction. Eberts et al [5] slice texts into segments and recognize entities for each segment. Yu et al [6] leverage the Biaffine attention mechanism on word-word matrices to allow information interaction between entity heads and tails. Different models address NER in unique settings and require specific input data formats.

Relation extraction aims to identify semantic relations between entity pairs. Recently, relation extraction methods are mainly based on deep learning. Perera et al. [7] propose a BiLSTM-CRF framework with attention for NER and RE on electronic medical records. Wu et al. [8] incorporate entity-level information for relation classification. There are also works focusing on joint extraction of entities and relations to avoid potential error propagation and missing task interconnections. To jointly decode entities and relations, Zheng et al. [9] propose a tagging-based method, Wang et al. [10] propose a form-filling method, and Zeng et al. [11] propose a sequence-to-sequence generation method. Similar to NER models, these methods require processing the raw data into specific formats. For example, the UniRE model [10] requires transforming the input sentence into an $n \times n$ matrix, where $n$ is the sentence length.

In knowledge extraction tasks, the diversity in datasets and model input formats compels researchers to craft custom data processing codes to convert datasets into model-specific formats. These codes are difficult to reuse and maintain, imposing a burden on model development and application.

### B. ModelOps

ModelOps is proposed to address the of lack of scientific management and automation in AI model development and deployment. Currently, notable platforms include MLFlow [12], MLReef [13], ModelScpoe [14], etc.

MLFlow provides experiment tracking, code packaging, model lifecycle management with lightweight APIs compatible with existing machine learning libraries. However, it lacks dataset retrieval, auto dataset adaptation, and mainly targets the needs of researchers and doesn't provide automated AI solutions for non-AI experts.

MLReef adopts a Git-like approach to manage machine learning projects but falls short in sharing models across projects. Additionally, manual dataset adaptation and computational resource allocation are needed, making the system less practical for non-AI experts.

ModelScope provides model services for users, offering a rich collection of pre-trained models, datasets, and the ability to perform model inference with a single line of code. It provides software development kits and a web interface to cater to different needs. Yet, the platform misses out model training, monitoring and doesn't support private web interface deployment. Integrating new models into the platform also involves a potentially cumbersome Pull Request (PR) process, which may hinder its practicability.

### C. Model Recommendation

Model recommendation is an important task in automated machine learning (Auto ML), and many studies have been conducted on this topic. For instance, Vainshtein et al [15] introduces the AutoDi method, Cohen-shapira et al proposes the TRIO method [16], and the AutoGRD method [17]. These approaches primarily rely on meta learning and focus on machine learning models and tabular datasets. Their fundamental assumption is that models trained on similar datasets exhibit similar performance.

In conclusion, although there are AI model management tools available, the support for knowledge extraction tasks is still limited. Furthermore, there are some major limitations in current technologies: insufficient support for dataset and model adaptation, inadequate monitoring during model experiments, and a lack of model recommendation methods for knowledge extraction.

## III. ARCHITECTURES

The proposed framework consists of a dataset abstraction mechanism and a reusable model training, monitoring and management mechanism for automated model training and evaluation. The overall architecture of the framework is shown in Figure 1. The data abstraction mechanism converts original datasets to model-specific formats, facilitating quick adaptation between the selected dataset and the model. The reusable model training, monitoring and management mechanism manages experiments, performs automated training and evaluation and monitors the entire lifecycle of the knowledge extraction model. These mechanisms will be detailed in the following parts.

### A. Dataset abstraction mechanism based on multi-layer callback functions

To address the challenges posed by the heterogeneity of knowledge extraction datasets and model-specific input requirements, we propose a dataset abstraction mechanism based on multi-layer callback functions. The mechanism offers flexibility catering to varied datasets and models.

As mentioned earlier, knowledge extraction datasets are inherently complex and heterogeneous, and models have unique input format requirements. For dataset-model adaptation, one-time codes are needed to convert the dataset into model-specific formats. For N datasets and M models, N×M pieces of codes are needed, and multiple adaptation codes are required when integrating new datasets and models. This complexity impedes development efficiency and hinders non-programmers to utilize knowledge extraction models.

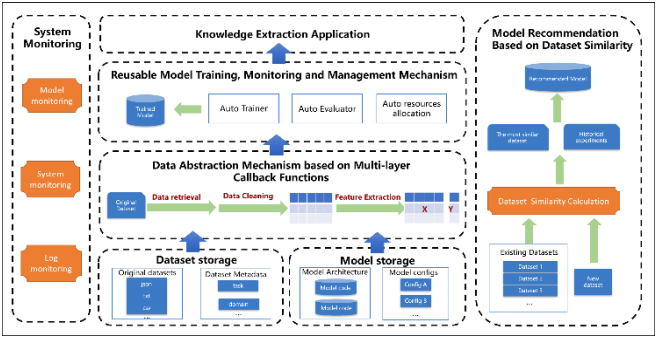

Figure 1: The overall architecture of the proposed framework

Therefore, our goal is to enable efficient, low-code dataset-model adaptation. Each dataset and model only needs one piece of processing codes upon integration into the platform and can then be used without additional glue codes. To this end, we design a unified data specification for every task. By encapsulating dataset files and corresponding processing methods into a dataset abstraction class, we mask the underlying details and present a standardized data format to the upper-level models. Model-specific processing codes can then be written solely based on this unified format. For $N$ datasets and $M$ models, only $N + M$ pieces of codes are needed, which highly reduces the coding redundancy.

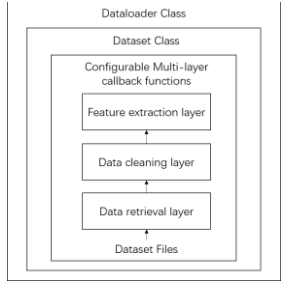

Figure 2: The design of our proposed multi-layer callback functions. By calling the three layers of callback functions, the original dataset will be converted to the model-specific format and used to construct a Dataloader class, facilitating quick adaptation between datasets and models.

To ensure the flexibility and scalability of our framework, we use multi-layer callback functions to realize the data loading process, as shown in Figure 2. We design three layers of callback functions: the data retrieval layer retrieves data and reads it into the standardized format, the data cleaning layer performs data cleaning, and the feature extraction layer processes standardized data into the model-specific input format. The three-layer design decouples data retrieval, cleaning and feature extraction, and callback functions allow dataset-specific retrieval functions and model-specific feature extraction functions to be passed as parameters and flexibly replaced and used as plugins, without modifying the code structure. In practice, when a dataset and a model are selected for training, their corresponding retrieval and feature extraction methods are automatically loaded through partial functions and callbacks, adapting the dataset to the model. During this process, no extra coding are needed, thus achieving low-code dataset-model adaptation.

### B. Reusable Model Training, Monitoring and Management Mechanism

When developing knowledge extraction models, in addition to designing the model architecture, it is also necessary to write codes for model training and evaluation, allocate computational resources, collect and analyze experimental results. Implementing these parts manually can be time-consuming and labor-intensive, and challenging for non-AI experts. Therefore, we design a model training, monitoring, and management mechanism to uniformly manage reusable components in this process. This mechanism automates resource allocation, model training, evaluation, and monitoring. When conducting model experiments, users only need to specify the dataset, the model and hyperparameters, while our framework handles the rest.

Our goal is automated model training and evaluation, enabling the platform to conduct experiments based on user-specified hyperparameters without manual code writing and other operations. To achieve this, we implement automatic allocation of computational resources and automated model training and evaluation.

For automatic resource allocation, we adopt a distributed architecture, to ensure horizontal scalability for potential increases in machine and GPU resources. The architecture comprises Service Gateway, AI Server, and Worker. Service Gateway acts as the entry point for model services, distributing tasks to available machines. AI Server runs on a machine, receiving requests and assigning tasks to Workers. It launches worker processes according to the number of available GPUs. Worker is bound to a GPU and executes assigned tasks. Service Gateway adopts a load-balancing algorithm based on weighted least connections, using the number of GPUs on each machine as weights and directing tasks to the chosen machine. AI Server and Workers communicate via multiprocessing queues. AI Server dispatches tasks in a round-robin way, writing task information to the message queue of the selected Worker. Worker waits in a blocking state and asynchronously executes tasks. Considering that the GPU memory required for a task is hard to predict, we specify that a worker executes only one task at a time to avoid memory overflow. Multiple tasks will be processed sequentially in a first-in first-out manner. Through the design of these three components, we realize the automatic computational resource allocation.

For automated model training and evaluation, we design the auto trainer and evaluator, allowing users to train models and calculate evaluation metrics without manually writing codes. We design a Trainer class to encapsulate training and evaluation, paired with a YAML-based configuration template. The designed workflow is as follows:

*1) Parse user configurations and generate a Config.*

*2) Load the specified model class according to the configuration, if a trained model is requried, load it from storage, otherwose initialize the model.*

*3) Execute the callback functions in the data abstraction mechanism to construct a DataLoader.*

*4) The Trainer class uses a reflection mechanism to check if the model class contains custom functions. If so, these functions are called to achieve customization in training. If not, default strategies are adopted.*

*5) Execute training, including forward passing, loss calculation, backpropagation and gradient updates. Evaluate the model using the evaluatorafter each epoch.*

For model evaluation, to ensure that the evaluator can be applied to different models, the models are required to implement a decoding method to produce standardized outputs. For NER, models need to output the boundary positions and types of entities; for RE, models need to output the relation types; for joint extraction, models need to output entity triplets and relation septuplets. Based on the unified output specifications, the auto evaluator calculates commonly used metrics, such as precision, recall, and F1 score.

After datasets and models are integrated into the framework, the aforementioned designs allow zero-code model training and evaluation by specifying the dataset and model.

We refer to the training and evaluation of a model as an experiment. Experiments are tracked from creation to completion, and relevant information such as model versions, hyperparameter lists, and metrics are recorded, which ensures reproducibility, and offers experiment comparisons for optimization. We also implement Git-based model version control, and comprehensive monitoring and visualization using TensorBoard, Prometheus, Loki and Grafana.

Through the above design, we realize a reusable model training, monitoring, and management mechanism, which enables efficient management of models and experiments.

## IV. MODEL RECOMMENDATION METHOD BASED ON DATASET SIMILARITY

During knowledge extraction model application, selecting a model from numerous options can be challenging. To aid users in selecting a suitable model for their dataset, we propose a model recommendation method based on dataset similarity.

Following current model recommendation literature, as introduced in related work, we assume models trained on similar datasets exhibit similar performance. Specifically, given dataset $d_i$ and model $m_i$, the performance is denoted as $f(d_i, m_i)$. Function $g$ converts the dataset into representations. When $g(d_1) \approx g(d_2)$. We assume that $f(g(d_1), m_1) \approx f(g(d_2), m_1)$.

Under the assumption, models performing well on similar datasets will also exhibit good performance on a new dataset. Therefore, our model recommendation method proceeds as: Firstly, categorize datasets by task and domain. Calculate dataset similarity within the category, and identify the existing dataset most similar to the given dataset. Next, gather the performance of models trained on this dataset, rank them based on desired performance metrics, and recommend the best-performing model. The core of this method is to calculate dataset similarity.

### A. Dataset Similarity calculation

Knowledge extraction datasets consist of textual data. To better capture semantic information of the text, we employ BERT-based pre-trained language models to convert texts into vector representations. We employ domain-specific BERT models for both training and dataset similarity calculation for consistent representations. To better reflect the overall distribution of the dataset and avoid information loss, we treat datasets as distributions and evaluate distributional similarities, instead of condensing them to a singular vector using techniques like averaging or concatenation. We utilize the following metrics to assess the similarity between two distributions:

**Maximum Mean Discrepancy (MMD)** measures the difference between two probability distributions, commonly used in transfer learning [18]. It maps the distributions into a space and calculates the maximum upper bound of the difference between the expectations of the mapped distributions. We map the original sample space to the Reproducing Kernel Hilbert Space (RKHS) and use kernel functions to do the calculation. Given two datasets generated from distributions $P$ and $Q$, with samples $x_1, x_2, \ldots, x_m$ and $y, y_2, \ldots, y_n$. The kernel-based MMD calculation is:

$$MMD^2(\mathcal{F}, P, Q) = \frac{1}{m^2} \sum_{i=1}^{m} \sum_{j=1}^{m} \kappa(x_i, x_j)$$
$$- \frac{2}{mn} \sum_{i=1}^{m} \sum_{j=1}^{m} \kappa(x_i, y_j) + \frac{1}{n^2} \sum_{i=1}^{m} \sum_{j=1}^{m} \kappa(y_i, y_j) \quad (1)$$

where $\kappa(\cdot, \cdot)$ is a kernel function. We use the Gaussian kernel function $\kappa(x, y) = exp(-\frac{\|x-y\|^2}{2\sigma^2})$. We average results from multiple Gaussian kernels to determine the final MMD metric between datasets.

**Fréchet Distance (FD)** measures the difference between two probability distributions. For Gaussian distributions x and y, FD is calculated as:

$$FD(\mathcal{F}, P, Q) = \|\mu x - \mu y\|_2^2$$
$$+ Tr(\Sigma_x + \Sigma_y - 2(\Sigma_x \Sigma_y)^{\frac{1}{2}}) \quad (2)$$

where $\mu x$ and $\mu y$ are mean vectors, $\Sigma_x$ and $\Sigma_y$ are covariance matrices of x and y. $Tr(\cdot)$ denotes the trace of a matrix and $\|\cdot\|^2$ denotes the 2-norm of a vector. Heusel et al [19] propose the Fréchet Inception Distance, leveraging the Inception network to extract image features to calculate the Fréchet Distance between generated and real images. In our scenario, for text datasets, we utilize the BERT model to obtain their features and then calculate the Fréchet Distance. This measure is referred to as Fréchet Bert Distance (FBD) and serves as an indicator to evaluate the similarity between text datasets.

Furthermore, considering that text datasets are long corpora, we follow the work by Kour et al. [20] and apply distributional text similarity metrics: PR distance and Mauve distance.

**The PR metric** by Sajjadi et al. [21] decomposes distributional differences into two independent dimensions: precision (sample similarity to the target distribution) and recall (sample coverage of target distribution diversity). F1 score is calculated as the final measurement.

**The Mauve metric** by Pillutla et al. [22] uses the Kullback-Leibler (KL) divergence in a quantized low-dimensional space to estimate the difference between text distributions.

After calculating metrics, we use rank-sum ratio for metric fusion. Datasets are ranked on each metric, and their weighted rank sums are computed and divided by the total sum of all objects to obtain the rank-sum ratio for each dataset. And the final similarity ranks can be acquired. This approach avoids potential distortion from relying on a single metric and helps mitigate the impact of different metric scales, making the fusion results more stable and comparable. Finally, based on similarity rankings, we identify the most similar existing dataset and recommend the best-performing model from past experiments on that dataset.

*B. experiments*

To validate the effectiveness of our model recommendation method, we conduct the following experiments: we train all candidate models on experimental datasets and obtain the optimal model for each dataset on each evaluation metric as the ground truth. We then compare our recommended model to the ground truth to evaluate the recommendation accuracy. A match with the optimal model is deemed correct and otherwise incorrect. We test four common NER models across seven datasets, with evaluation results shown in Table I. In Table I, P represents precision, R represents recall, and F represents F1 score. Based on these results, we obtain the optimal model for each dataset on each metric as gold recommendations. For example, the bert-span model has the highest F1 score on the bank dataset.

Considering that our experimental datasets do not have obvious domain characteristics, we use the universal bert-base-chinese model to convert datasets into representations. When calculating MMD, due to GPU constraints, we randomly sample 1000 texts from the datasets for calculation and take the average of ten calculations as the final result. The computed MMD values are shown in Table II. The diagonal values in Table II reflect the distribution differences between two subsets randomly extracted from the same original dataset. The rather small MMD values indicate small distributional differences between subsets of a dataset, which demonstrates the validness of our similarity calculation method. Similarly, we calculate FBD, PR and Mauve distances between each pair of datasets. Then, we obtained the recommended models via the proposed recommendation method. We select precision, recall, and F1 score as the desired metrics and compare recommended models to the ground truth. The results are shown in Table III.

TABLE I. MODEL EVALUATION RESULTS

| Datasets | | Models | | | |
|---|---|---|---|---|---|
| | | Bert_softmax | Bert_bilstm_crf | Bert_crf | Bert_span |
| Bank | P | 0.8602 | 0.8453 | 0.8501 | 0.9622 |
| | R | 0.8368 | 0.8555 | 0.8434 | 0.8064 |
| | F | 0.8483 | 0.8504 | 0.8467 | 0.8774 |
| Ecommerce | P | 0.8532 | 0.8810 | 0.8547 | 0.9597 |
| | R | 0.8617 | 0.8878 | 0.8663 | 0.8067 |
| | F | 0.8483 | 0.8845 | 0.8604 | 0.8766 |
| Finance | P | 0.8700 | 0.8615 | 0.8797 | 0.9020 |
| | R | 0.8836 | 0.8805 | 0.8742 | 0.8679 |
| | F | 0.8768 | 0.8709 | 0.8770 | 0.8846 |
| Resume | P | 0.9571 | 0.9510 | 0.9511 | 0.9598 |
| | R | 0.9600 | 0.9534 | 0.9540 | 0.9515 |
| | F | 0.9587 | 0.9522 | 0.9525 | 0.9556 |
| Weibo | P | 0.6667 | 0.6533 | 0.6726 | 0.7432 |
| | R | 0.6584 | 0.5644 | 0.6510 | 0.6733 |
| | F | 0.6625 | 0.6056 | 0.6616 | 0.7065 |
| Renmin | P | 0.9628 | 0.9566 | 0.9622 | 0.9683 |
| | R | 0.8080 | 0.8020 | 0.8064 | 0.8056 |
| | F | 0.8787 | 0.8725 | 0.8774 | 0.8795 |
| Nlpcc | P | 0.8912 | 0.9097 | 0.9117 | 0.9563 |
| | R | 0.8979 | 0.9121 | 0.9071 | 0.8056 |
| | V | 0.8945 | 0.9110 | 0.9094 | 0.8745 |

TABLE II. RESULTS OF DATASET MMD DISTANCE

| datasets | datasets | | | | | | |
|---|---|---|---|---|---|---|---|
| | bank | resume | weibo | finance | renmin | ecomm | nlpcc |
| Bank | 0.0058 | 1.2447 | 0.4132 | 0.5932 | 0.6510 | 0.7311 | 0.5458 |
| resume | 1.2447 | 0.0044 | 1.1211 | 0.9534 | 0.9489 | 1.1102 | 1.1435 |
| Weibo | 0.4132 | 1.1211 | 0.0017 | 0.4432 | 0.4259 | 0.5030 | 0.5402 |
| Finance | 0.5932 | 0.9534 | 0.4432 | 0.0013 | 0.0914 | 0.6373 | 0.7028 |
| Renimn | 0.6510 | 0.9489 | 0.4259 | 0.0914 | 0.0057 | 0.5712 | 0.6697 |
| Ecomm, | 0.7311 | 1.1102 | 0.5030 | 0.6373 | 0.5712 | 0.0049 | 0.5297 |
| Nlpcc | 0.5458 | 1.1435 | 0.5402 | 0.7028 | 0.6697 | 0.5297 | 0.0055 |

TABLE III. RESULTS OF RECOMMENDATION ACCURACY

| Similarity Metrics | Accuracy |
|---|---|
| MMD-based | 0.76 |
| FBD-based | 0.71 |
| PR-based | 0.76 |
| Mauve-based | 0.76 |
| Rank-sum ratio | 0.81 |

Table III shows the accuracy of model recommendations using different similarity metrics. MMD, FBD, PR, and mauve-based rankings have accuracies of 0.76, 0.71, 0.76, and 0.76 respectively. The rank-sum ratio method, combining multiple metrics, achieves the highest accuracy at 0.81, making it the most effective recommendation method. While our experiments are conducted on general datasets, we believe that our method is universally applicable and its conclusions remain valid in the medical domain

In this section, we conduct experiments on four models and seven datasets to validate the effectiveness of our proposed model recommendation method. The experimental results show that the rank-sum ratio method, fusing multiple metrics, achieves an accuracy of 0.81. This method can provide users with a potentially well-performing model on their target datasets, thus reducing time and computational resources in exploring different model architectures.

## V. SYSTEM IMPLEMENTATIONS

Based on the proposed framework, we design and implement a ModelOps-based medical knowledge extraction platform. The platform features a user-interface frontend, a backend for database and model service access, and a model server providing model training, evaluation, prediction, monitoring and other abilities. We adopt a frontend-backend separation pattern. We develop the frontend using Vue.js and the backend using Django REST framework. The frontend and the backend communicate via HTTP, and the backend interacts with the model server via gRPC.

**Frontend Website.** Provides five main pages: Dataset, Model, Experiment, Monitoring, and Documentation. Users can view, search, and download datasets; search and retrieve model details and codes; create and track experiments; monitor the system; and access platform documentation. The user interface for experiment creation is shown in figure 3.

**Platform Backend** Handles the logic of the frontend website and acts as a proxy to access model services. It maintains metadata, manages file storage and model Git repositories, and monitors experiments.

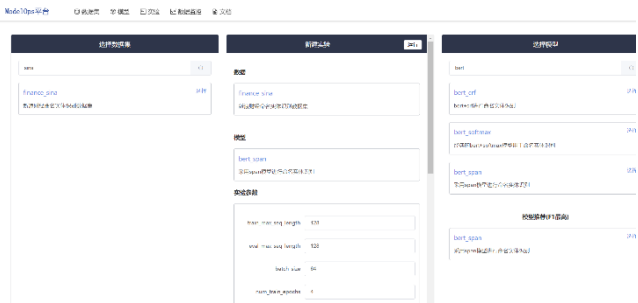

Figure 3: The user interface for experiment creation. Users can start model training and evaluation simply by selecting a dataset, a model and configuring hyperparameters, without writing a single line of code.

**Model Server**. Provides various model-related abilities as services (Model as a Service, MaaS), receives tasks and performs continuous monitoring.

The main workflow of the platform is: After selecting a dataset, the platform recommends suitable models. Once a model is selected, the platform starts an experiment, automatically handles resource allocation, model training and evaluation, and returns a ready-to-use model.

Integrating the above three components, the ModelOps platform offers an all-in-one solution for knowledge extraction. For researchers, the platform improves efficiency and allows them to focus on model design. For non-technical users, like doctors, the platform offers low-code model construction and application.

## VI. CONCLUSION

In this paper, we introduce a ModelOps-based framework for intelligent medical knowledge extraction. The framework features a dataset abstraction mechanism based on multi-layer callback functions, a reusable model training, monitoring and management mechanism, and a model recommendation method based on dataset similarity. Based on the proposed methods, we design and implement a modelOps-based knowledge extraction system. The system addresses the challenges in knowledge extraction model construction and application, providing convenience for researchers to develop models and non-AI expert users such as doctors to apply models in real scenarios.

## ACKNOWLEDGMENT

This work was supported by the National Natural Science Foundation of China (Grant No.62172011) and the Fundamental Research Funds for the Central Universities of Ministry of Education of China.